\documentclass[11pt]{article}

\usepackage[preprint]{acl}

\usepackage{times}
\usepackage{latexsym}

\usepackage[T1]{fontenc}
\usepackage[utf8]{inputenc}

\usepackage{microtype}

\usepackage{inconsolata}

\usepackage{graphicx}
\usepackage{amsmath}
\usepackage{booktabs}
\usepackage{multirow}
\usepackage{xcolor}
\usepackage{array}
\definecolor{right}{HTML}{97D077}
\definecolor{wrong}{HTML}{EA6B66}
\usepackage{tabularx}
\usepackage{enumitem}

\title{Pardon? Evaluating Conversational Repair \\in Large Audio-Language Models}

\author{
 \textbf{Shuanghong Huang\textsuperscript{1}\textsuperscript{$*$}},
 \textbf{Jinlei Xu\textsuperscript{1}\thanks{Equal contribution.}},
 \textbf{Youchao Zhou\textsuperscript{1}},\\
 \textbf{Yanghao Zhou\textsuperscript{1}},
 \textbf{Xuan Zhao\textsuperscript{1}},
 \textbf{Chong Feng\textsuperscript{1}\textsuperscript{$\dagger$}},
 \textbf{Wenxuan Zhang\textsuperscript{2}\thanks{Corresponding author.}}
\\
\\
 \textsuperscript{1}Beijing Institute of Technology,
 \textsuperscript{2}Singapore University of Technology and Design
\\
 \small{
   \{shuanghong, xujinlei, fengchong\}@bit.edu.cn, wxzhang@sutd.edu.sg
 }
}

\begin{document}
\maketitle
\begin{abstract}
Large Audio-Language Models (LALMs) have demonstrated strong performance in spoken question answering (QA), with existing evaluations primarily focusing on answer accuracy and robustness to acoustic perturbations. However, such evaluations implicitly assume that spoken inputs remain semantically answerable, an assumption that often fails in real-world interaction when essential information is missing.
In this work, we introduce a repair-aware evaluation setting that explicitly distinguishes between answerable and unanswerable audio inputs. We define answerability as a property of the input itself and construct paired evaluation conditions using a semantic-acoustic masking protocol. Based on this setting, we propose the Evaluability Awareness and Repair (EAR) score, a non-compensatory metric that jointly evaluates task competence under answerable conditions and repair behavior under unanswerable conditions.
Experiments on two spoken QA benchmarks across diverse LALMs reveal a consistent gap between answer accuracy and conversational reliability: while many models perform well when inputs are answerable, most fail to recognize semantic unanswerability and initiate appropriate conversational repair. These findings expose a limitation of prevailing accuracy-centric evaluation practices and motivate reliability assessments that treat unanswerable inputs as cues for repair and continued interaction. The core code and dataset are publicly available at
\url{https://github.com/sheunghung/EAR}.
\end{abstract}

\section{Introduction}
Large Audio-Language Models~\cite{cui2025recent,arora2025on} (LALMs) have rapidly evolved from passive speech recognition systems into interactive conversational agents capable of reasoning and decision-making directly from spoken input~\cite{tang2024salmonn, xu2025qwen2}. Recent models such as GPT-4o \cite{hurst2024gpt}, Gemini~2.5~\cite{comanici2025gemini} and DeSTA2.5-Audio \cite{lu2025desta2} demonstrate strong performance on spoken question answering~\cite{gong2024listen} (QA) and related tasks, enabling a wide range of applications including virtual assistants~\cite{anastassiou2024seed}, real-time translation~\cite{barrault2023seamlessm4t}, and multimodal dialogue systems~\cite{fang2025llamaomni2}. As LALMs increasingly operate in open-world conversational settings, their reliability hinges not only on recognizing speech accurately but also on responding appropriately to varying input conditions.

\begin{figure*}[t]
  \centering
  \includegraphics[width=.9\textwidth]{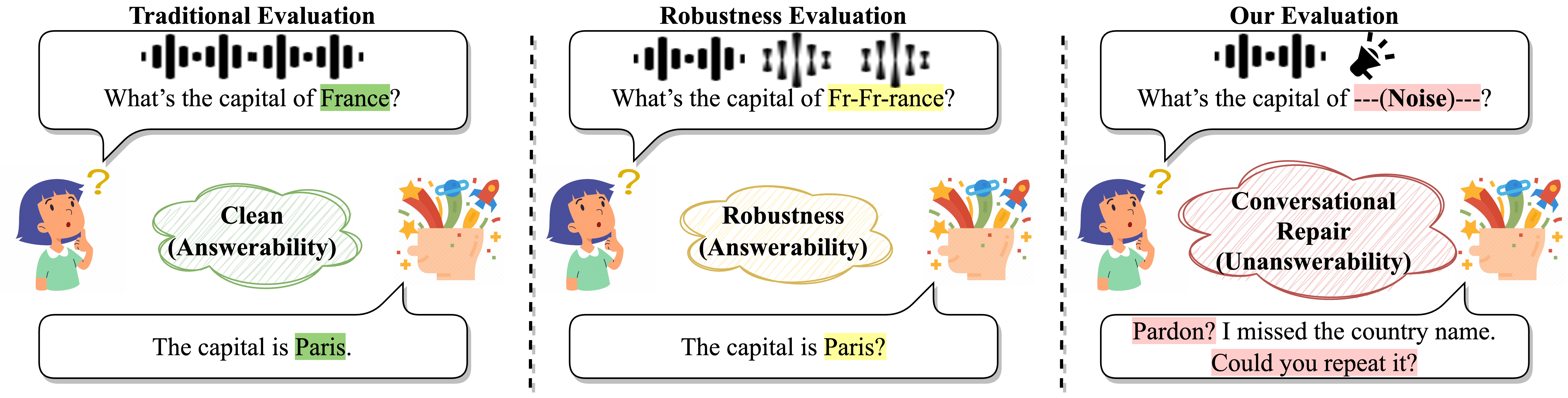}
  \caption{\textbf{Left}: Traditional evaluation measures answer correctness under clean, answerable inputs. \textbf{Middle}: Robustness evaluation tests whether correct answers are maintained under acoustic perturbations that preserve answerability. \textbf{Right}: Our repair-aware evaluation masks answer-critical information to create unanswerable inputs and assesses whether models shift from answering to conversational repair.}
  \label{fig_intro}
\end{figure*}

As illustrated in Figure~\ref{fig_intro}, typical evaluations of LALMs primarily assess performance through answer accuracy under clean speech inputs, focusing on whether models can produce the correct response to a given query~\cite{lipping2022clotho, yang2024air, sakshi2025mmau}. To further test their reliability, more recent benchmarks extend evaluation to robustness under adverse acoustic conditions, such as noise, disfluency, or signal corruption~\cite{ma2025towards, liu2025vocalbench}. In these settings, degraded audio is treated as a more challenging but still semantically answerable variant of the original input, and model outputs are evaluated against the fixed ground-truth answer.

This robustness-centered paradigm has been effective for measuring speech understanding under adverse conditions. However, it implicitly assumes a crucial premise: that the input remains semantically \textbf{\textit{answerable}}. In real-world spoken communication, this assumption often breaks down. Audio inputs may become incomplete due to noise, transmission loss, or overlapping speech, leading to the loss of answer-critical semantic information. In such cases, the problem extends beyond reduced accuracy, as the input becomes intrinsically \textbf{\textit{unanswerable}} even for a human listener. Rather than guessing or disengaging, human interlocutors respond through conversational repair, signaling the breakdown (e.g., “Pardon?”) and requesting clarification to address the issue. Reliable spoken interaction, therefore, depends on the ability to recognize when semantic evaluability is lost and to adapt conversational behavior accordingly.

However, prevailing evaluations for LALMs remain answer-centric, treating semantically incomplete inputs as if they required a definitive answer. Under this framing, models are encouraged to hallucinate plausible-looking responses or issue generic refusals, rather than adapting their behavior through conversational repair. Such outputs obscure whether the model has recognized the loss of semantic evaluability and fail to support conversational recovery.

In this work, we introduce a repair-aware evaluation setting that makes answerability an explicit part of the evaluation design. We frame answerability as a property of the audio input itself and separate evaluation into answerable and unanswerable conditions. Unanswerable inputs are defined as those in which answer-critical semantic information is absent, such that even a human listener cannot determine the correct answer.

Under this setting, reliable conversational behavior requires models to condition their responses on semantic evaluability: producing task-fulfilling answers when inputs are answerable, and initiating appropriate conversational repair when they are not. To operationalize this distinction, we propose a semantic-acoustic masking protocol that selectively removes answer-critical semantic content, yielding paired answerable and unanswerable inputs derived from the same underlying query. Building on this controlled setup, we introduce the Evaluability Awareness and Repair (EAR) score, a non-compensatory metric that jointly evaluates task competence under answerable conditions and repair behavior under unanswerable conditions.

Experiments on two spoken QA benchmarks across a diverse set of LALMs reveal a consistent gap between answer accuracy and conversational reliability. While many models achieve high accuracy when inputs are answerable, most fail to recognize when essential semantic information is missing and do not initiate conversational repair. These results indicate that strong robustness or accuracy alone does not imply evaluability awareness, and that prevailing evaluation practices substantially overestimate real-world conversational reliability.

Our contributions are summarized as follows:
\begin{itemize}[noitemsep, topsep=0pt, left=0pt]
    \item We identify that existing LALM evaluations assess answer accuracy and robustness, but omit conversational repair behavior.
    \item We introduce a repair-aware evaluation setting that explicitly distinguishes answerable and unanswerable spoken inputs.
    \item We propose a semantic-acoustic masking protocol and the EAR score to jointly evaluate task competence and conversational repair without mutual compensation.
\end{itemize}

\section{Related Work}
\subsection{Large Audio-Language Models}
LALMs extend multimodal large language models (LLMs) to spoken inputs, enabling instruction following and language-style reasoning grounded in audio. Compared with text, audio signals exhibit greater heterogeneity across speech, music, and environmental sounds, involving diverse temporal structures and acoustic patterns. To accommodate such variability, LALMs are designed to support a unified spoken interface that can flexibly process a wide range of audio inputs within a single model, rather than relying on task-specific audio-language pipelines~\cite{cui2025recent,arora2025on}.

Most LALMs adopt a modular architecture that integrates a pretrained audio encoder with an LLM backbone, allowing auditory representations to be aligned with language generation and reasoning. Recent work has increasingly focused on \emph{general-purpose} LALMs that support audio instruction following and broad auditory reasoning across domains, moving beyond narrowly scoped tasks such as speech recognition or audio classification. Representative systems including GAMA~\cite{ghosh2024gama}, SALMONN~\cite{tang2024salmonn}, Qwen2-Audio~\cite{chu2024qwen2}, and DeSTA2.5-Audio~\cite{lu2025desta2} demonstrate steady progress toward more versatile spoken interaction capabilities within a unified framework.

\subsection{Evaluating Reliability and Repair}
Evaluating the reliability of language models has received increasing attention in recent years, although a unified standard remains elusive. Prior work has proposed a range of metrics and datasets to encourage models to abstain when reliable answers cannot be produced, capturing notions such as prudence, honesty, and truthfulness~\cite{yang2024alignment,cheng2024can}, as well as faithfulness- and precision-based measures~\cite{yona2024can,zhang-etal-2024-r}. More recent approaches further integrate answer accuracy and refusal behavior into unified reliability scores by explicitly modeling their trade-offs~\cite{xu2024rejection}. Despite these advances, such reliability evaluations predominantly operate on static textual inputs and treat reliability as a property of isolated input-output pairs.

In spoken and conversational settings, reliability manifests differently from text-based scenarios. Rather than reflecting knowledge uncertainty, failures often arise from audio-level degradation, where answer-critical semantic information is partially or entirely missing from the speech signal. Current benchmarks for LALMs primarily emphasize robustness, assessing whether task performance can be maintained under noisy or disfluent speech conditions~\cite{ma2025c3,chen2024voicebench}. Extensions such as VOCALBENCH-DF~\cite{liu2025vocalbench} further analyze model behavior under diverse speech disfluencies, but generally assume that sufficient answer-critical semantic information is preserved in the audio, focusing on robustness rather than semantic unanswerability caused by audio-level information loss.

In contrast to prior work, our evaluation explicitly distinguishes between answerable and unanswerable conversational states derived from the same underlying query. Moreover, rather than treating refusal as the sole reliable fallback, we conceptualize conversational repair as a constructive and measurable behavior, evaluating whether models appropriately shift from answering to clarification when semantic evaluability is lost.

\begin{figure*}[t]
    \centering
    \includegraphics[width=.8\linewidth]{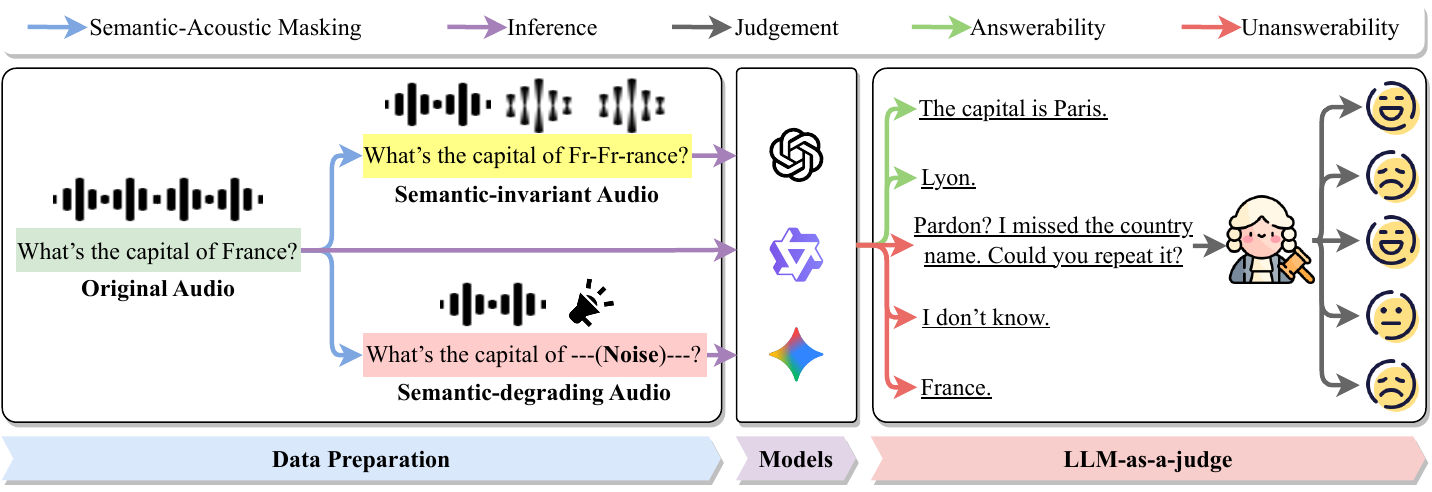}
    \caption{Overview of the repair-aware evaluation framework. The original audio generates two variants using the semantic-acoustic masking protocol: an answerable (semantic-invariant) and an unanswerable (semantic-degrading). The same set of LALMs then processes the original audio and its two masked variants. Finally, the LLM-as-a-judge evaluates these responses for answer correctness and conversational repair, respectively, under answerable and unanswerable inputs.}
    \label{fig_framework}
\end{figure*}

\section{Methodology}
To evaluate conversational reliability under varying semantic conditions, we propose a repair-aware evaluation framework, illustrated in Figure~\ref{fig_framework}. The core idea is to explicitly control the answerability of spoken inputs while maintaining the underlying query and target answer. Starting from an original audio, we generate semantically distinct audio variants through the semantic-acoustic masking protocol, and assess whether models appropriately condition their behavior on input evaluability, answering when sufficient information is available and initiating conversational repair when it is not.

\subsection{Problem Definition}
We study the evaluation of LALMs in spoken QA, focusing on whether models can adapt their conversational behavior to the semantic evaluability of spoken inputs. Let~$q$ denote an underlying semantic query, associated with a correct answer~$y$. The query~$q$ is modality-independent and specifies the information required at the semantic level. Given a spoken QA instance corresponding to~$q$, we construct multiple audio realizations~$x$ by applying controlled acoustic masking, resulting in inputs that differ in semantic completeness while sharing the same underlying query. We define an answerability function:
\begin{equation}
    A(x) \in \{0,1\},
\end{equation}
where~$A(x)=1$ indicates that the audio input~$x$ contains sufficient semantic information for a human listener to determine~$y$, and~$A(x)=0$ indicates that essential semantic information is missing. Answerability is thus defined as a property of the audio input itself, independent of model behavior.

For each semantic query~$q$, we construct two evaluation conditions: an answerable audio input~$x^{a}$ with~$A(x^{a})=1$, and an unanswerable audio input~$x^{u}$ with~$A(x^{u})=0$. Both are derived from the same underlying query~$q$ and are associated with the same target answer~$y$. This paired design isolates the effect of semantic evaluability while controlling for task definition and label identity.

Given a model~$f$, reliable conversational behavior is defined functionally. Under the answerable condition~$x^{a}$, the model should produce a task-fulfilling answer consistent with~$y$. Under the unanswerable condition~$x^{u}$, behavior instead requires conversational repair, where the model explicitly recognizes missing information and requests clarification, rather than attempting to answer.

\subsection{Semantic-Acoustic Masking Protocol}
To operationalize the answerability function $A(x)$ over audio inputs, we introduce a semantic-acoustic masking protocol that constructs paired answerable and unanswerable audio variants by selectively masking answer-critical or non-critical semantic content.

At a conceptual level, we define a semantic unit as answer-critical if its removal renders the spoken input underspecified for a human listener, such that the question can no longer be reliably answered. In practice, identifying minimal answer-critical units in arbitrary speech is challenging. To enable controlled and reproducible masking, we adopt a simplified operational strategy that leverages the availability of ground-truth answers and automatic speech processing tools. Across all conditions, masking is applied at the acoustic level by aligning text tokens to word-level time spans and replacing the corresponding segments with controlled signals (e.g., silence or white noise), while preserving the overall temporal structure of the utterance.

\paragraph{Semantic-degrading masking (unanswerable condition).}
For constructing semantically unanswerable inputs, we directly mask the audio segments corresponding to the ground-truth answer. Specifically, given an audio instance with a known answer span, we align the answer text to the audio and replace the corresponding time segments with controlled acoustic signals. This procedure ensures that answer-critical semantic information is absent from the audio, making the input intrinsically unanswerable even for a human listener ($A(x)=0$).

\paragraph{Semantic-invariant masking (answerable condition).}
For constructing answerable but acoustically perturbed inputs, we aim to introduce surface-level acoustic variation without affecting semantic evaluability. We first obtain an automatic transcription of the audio using Whisper-large-v3\footnote{https://huggingface.co/openai/whisper-large-v3}~\cite{radford2023robust} and perform part-of-speech tagging on the transcript using spaCy\footnote{https://github.com/explosion/spaCy}. We then randomly select a single token from a predefined set of function-word categories (e.g., determiners, adpositions, auxiliaries, conjunctions, pronouns, and particles) and replace its corresponding audio segment with controlled acoustic signals. Because such function words do not carry answer-critical semantic information, this masking preserves answerability while altering the acoustic surface ($A(x)=1$).

\subsection{Evaluation Triplets and Metrics}
From each underlying semantic query $q$, we construct an evaluation triplet~$(x^{a}, x^{u}, y)$. The answerable condition~$x^{a}$ includes multiple semantically equivalent realizations, specifically the original audio and its semantic-invariant masked variant. The unanswerable condition~$x^{u}$ is instantiated via semantic-degrading masking.

\paragraph{Task Competence under Answerable Conditions.}
Task competence is evaluated only when~$A(x)=1$. We define
\begin{equation}
    C(x^{a}) =
        \begin{cases}
            1, & \text{if answers correctly},\\
            0, & \text{otherwise}.
        \end{cases}
\end{equation}
In practice,~$C(x^{a})$ is computed as the average accuracy over all answerable realizations of~$x^{a}$, including both the original audio and the semantic-invariant masked input. This captures whether the model can reliably perform the task when sufficient semantic information is available, despite surface-level acoustic variation.

\paragraph{Repair Behavior under Unanswerable Conditions.}
Under the unanswerable condition~$x^{u}$, correctness is ill-defined. Instead, we assess the model’s conversational repair behavior via a graded score:
\begin{equation}
    R(x^{u}) =
        \begin{cases}
            1, & \text{explicit repair},\\
            0.5, & \text{generic refusal},\\
            0, & \text{otherwise}.
        \end{cases}
\end{equation}
Explicit repair refers to responses that identify missing information and request targeted clarification. Generic refusals (e.g., “I don’t know”) receive partial credit, while hallucinated or incorrect answers receive none.

\paragraph{Evaluability Awareness and Repair Score}
To jointly evaluate conditional conversational reliability, we define the EAR score as a dataset-level metric. Let~$C$ denote the average task competence over all answerable audio inputs, and let~$R$ denote the average repair behavior score over all unanswerable inputs across the dataset. The EAR score is defined as the harmonic mean of~$C$ and~$R$:
\begin{equation}
    \mathrm{EAR} = \frac{2 \cdot C \cdot R}{C + R}.
\end{equation}
This non-compensatory formulation ensures that high scores are achieved only when a model both answers correctly under answerable conditions and performs repair under unanswerable conditions.

\paragraph{Repair Behavior Identification}
Repair behavior is identified using an LLM-as-a-Judge paradigm~\cite{li2025generation}. Given the query, the masked input context, and the model response, a judge model is instructed to categorize the response into explicit repair, generic refusal, or other. The judge produces categorical decisions only, reducing subjectivity and improving reproducibility. This procedure is used solely for evaluation and does not provide supervision to the evaluated models.

\begin{table*}[t]
\centering
\small
\setlength{\tabcolsep}{12pt}
\renewcommand{\arraystretch}{1.2}
\begin{tabular}{cccccccc}
\toprule
\multirow{2}{*}{\centering\textbf{Model}} 
& \multicolumn{3}{c}{\textbf{WDYL}} 
& \multicolumn{3}{c}{\textbf{SLUE-SQA-5}} \\
\cmidrule(lr){2-4} \cmidrule(lr){5-7}
& \textbf{$C$~$\uparrow$} 
& \textbf{$R$~$\uparrow$} 
& \textbf{EAR~$\uparrow$} 
& \textbf{$C$~$\uparrow$} 
& \textbf{$R$~$\uparrow$} 
& \textbf{EAR~$\uparrow$}   \\
\midrule
Qwen2-Audio~\cite{chu2024qwen2} 
& 5.5 & 37.7 & 9.6 
& 29.1 & 10.1 & 15.0 \\
Baichuan-Omni~\cite{li2024baichuan} 
& 70.3 & 3.1 & 5.0 
& 51.5 & 3.0 & 5.7 \\
Qwen2.5-Omni~\cite{xu2025qwen2} 
& 77.3 & 1.0 & 2.0 
& 53.5 & 5.4 & 9.8 \\
DeSTA2.5-Audio~\cite{lu2025desta2} 
& 70.8 & 14.2 & \underline{23.7}
& 39.5 & 27.7 & \underline{32.6} \\
Audio Flamingo~3~\cite{ghosh2025audio} 
& 82.2 & 6.2 & 11.5 
& 40.6 & 11.4 & 17.8 \\
GPT-4o~\cite{hurst2024gpt} 
& 91.5 & 25.4 & 39.8 
& 43.8 & 36.2 & \textbf{39.6} \\
Gemini~2.5~\cite{comanici2025gemini} 
& 99.5 & 63.0 & \textbf{77.2} 
& 65.7 & 11.2 & 19.1 \\
\bottomrule
\end{tabular}
\caption{Repair-aware evaluation results across two spoken QA datasets. We report task competence under answerable inputs ($C$), repair behavior under unanswerable inputs ($R$), and the resulting EAR score. \textbf{Bold} denotes the best overall performance, and \underline{underlined} indicates the best open-source model.}
\label{tab_main}
\end{table*}

\section{Experimental Setup}
\subsection{Datasets}
We evaluate our framework on two spoken QA benchmarks with complementary conversational and acoustic characteristics, enabling assessment of repair-aware behavior under both interaction-driven and acoustically challenging conditions.

\paragraph{What Do You Like? (WDYL).}
WDYL~\cite{wu2024just} contains 1,000 spoken questions paired with audio recordings and ground-truth answers. The dataset features conversational, interaction-oriented queries that are often context-dependent, clarifying a natural and appropriate response when critical information is missing.

\paragraph{SLUE-SQA-5.}
SLUE-SQA-5~\cite{shon2023slue} is a large-scale spoken QA benchmark with longer utterances and more challenging acoustic conditions. We randomly sample 1,000 instances from the original dataset for evaluation. Compared to WDYL, answers in SLUE-SQA-5 are typically distributed over extended audio segments, increasing the impact of masking answer-critical content and providing a complementary testbed for repair-aware evaluation.

For each dataset, we construct evaluation triplets~$(x^{a}, x^{u}, y)$ using the semantic-acoustic masking protocol. Answerable inputs~$x^{a}$ retain sufficient information for answering, while unanswerable inputs~$x^{u}$ are rendered underspecified by masking answer-critical segments, as verified by human listeners.

\subsection{Models for Evaluation}
We evaluate a total of seven LALMs, covering both open-source and closed-source systems, to assess whether repair-aware conversational behavior is consistently exhibited across model families. The evaluated models include the open-source Qwen2-Audio~\cite{chu2024qwen2}, Baichuan-Omni~\cite{li2024baichuan}, Qwen2.5-Omni~\cite{xu2025qwen2}, DeSTA2.5-Audio~\cite{lu2025desta2}, and Audio Flamingo~3~\cite{ghosh2025audio}, as well as the closed-source GPT-4o~\cite{hurst2024gpt} and Gemini~2.5~\cite{comanici2025gemini}.

Open-source models are evaluated using their publicly released checkpoints and default inference configurations. Closed-source models are accessed through official APIs following recommended usage guidelines. All models are evaluated in a zero-shot setting, without task-specific fine-tuning, which reflects deployment scenarios and avoids confounding effects from supervision.

\subsection{Inference and Evaluation}
All experiments are conducted on NVIDIA RTX~3090 and A6000 GPUs. For open-source models, inference is performed using standard decoding settings commonly adopted in prior work, including greedy decoding or low-temperature sampling. Specifically, we use a temperature of 0 or 0.1 (when sampling is enabled), and set the maximum generation length to 2048 tokens, which is sufficient to cover both task answers and clarification-oriented repair responses. Closed-source models are queried using the default decoding parameters provided by their respective APIs.

For each evaluation triplet~$(x^{a}, x^{u}, y)$, each model is queried independently under the answerable and unanswerable conditions, producing responses~$r^{a}$ and~$r^{u}$, respectively. Answer correctness under the answerable condition is determined by exact match or semantic equivalence with the ground-truth answer~$y$, following standard practices in spoken QA. Repair behavior under the unanswerable condition is identified using an LLM-as-a-judge protocol instantiated with GPT-4o (gpt-4o-2024-11-20\footnote{https://platform.openai.com/docs/models/gpt-4o})~\cite{hurst2024gpt}. Given the original query, the masked input context, and the model response, GPT-4o is instructed to perform a categorical judgment of the response type (explicit repair, generic refusal, or otherwise).

We report the EAR score as the primary metric for conversational reliability. To make the computation of EAR transparent, we also report its constituent dataset-level components: task competence under answerable conditions, measured as the average answer accuracy over all answerable inputs (which defines~$C$), and repair performance under semantically degrading perturbations, measured as the average repair behavior score over all unanswerable inputs (which defines~$R$). All evaluations are conducted using a unified pipeline to ensure consistency across models and datasets.

The prompt templates used for model inference and evaluation are provided in Appendix~\ref{app_prompts}.

\begin{figure*}[t]
    \centering
    \includegraphics[width=\linewidth]{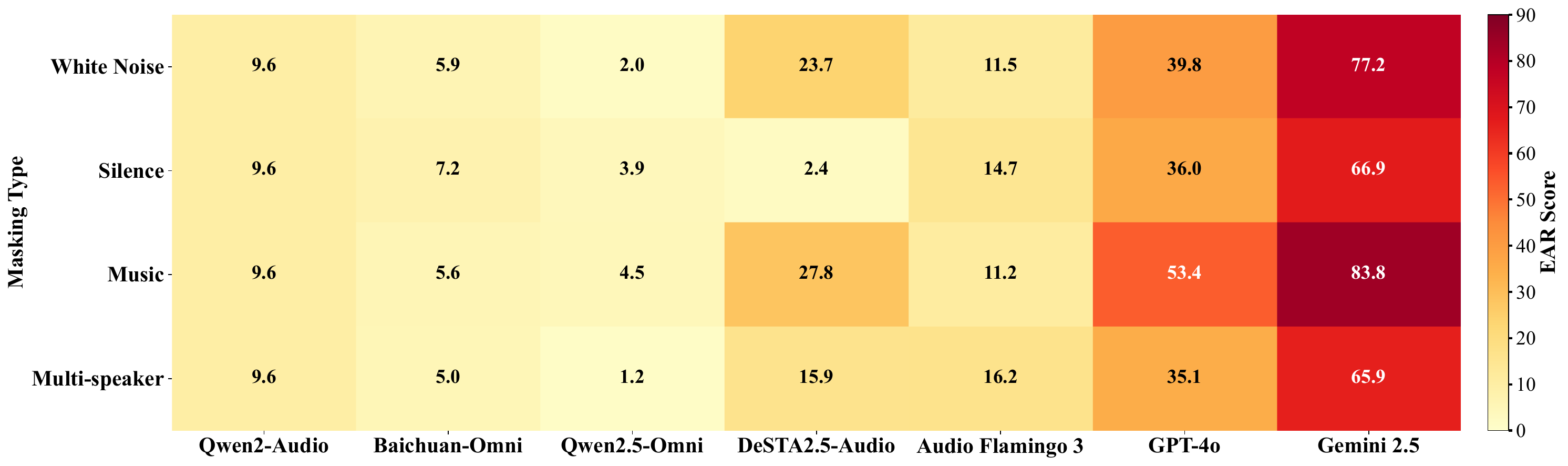}
    \caption{Sensitivity of EAR to different semantic-degrading masking realizations on the WDYL dataset. EAR scores are reported under four masking types: white noise, silence, music, and multi-speaker. While absolute values vary, relative model ordering is preserved across masking realizations.}
    \label{fig_stable}
\end{figure*}

\section{Experimental Results and Analysis}
\subsection{Main Results}
Table~\ref{tab_main} summarizes the repair-aware evaluation results across two spoken QA datasets. We report task competence under answerable conditions ($C$), repair behavior under unanswerable conditions ($R$), and the resulting EAR score, which jointly reflect conditional conversational reliability.

\paragraph{Finding 1: High task competence does not imply reliable conversational behavior.}
Across both datasets, many models achieve strong competence under answerable inputs, with~$C$ often exceeding 70\% on WDYL and remaining moderately high on SLUE-SQA-5. However, this competence does not consistently translate into high EAR scores. The primary reason is limited repair behavior: models that answer accurately under answerable conditions frequently fail to initiate repair once answer-critical information is removed. This pattern is especially evident for models such as Baichuan-Omni and Qwen2.5-Omni, which exhibit high~$C$ but consistently low~$R$ (typically remaining in the low single-digit range), resulting in substantially reduced EAR. These results indicate that accuracy-oriented evaluation can substantially overestimate model reliability by ignoring behavior under semantic failure.

\paragraph{Finding 2: Repair, rather than robustness, is the dominant bottleneck.}
Many models maintain moderately high~$C$ across datasets, suggesting reasonable robustness to content-preserving perturbations. In contrast,~$R$ varies dramatically and is the primary driver of EAR differences. Models that demonstrate explicit repair behavior, such as DeSTA2.5-Audio and Audio Flamingo~3, achieve substantially higher EAR despite having similar or even lower~$C$ values than other models. This separation confirms that repair awareness captures a distinct behavioral capability that cannot be explained by robustness or answer accuracy alone.

\paragraph{Finding 3: Recognizing semantic unanswerability remains challenging.}
Repair-aware reliability differs markedly between datasets. Compared to SLUE-SQA-5, models exhibit higher~$R$ and EAR scores on WDYL, which features shorter and more interaction-oriented queries. In contrast,~$R$ on SLUE-SQA-5 remain low even for strong closed-source models such as Gemini~2.5, indicating that longer and acoustically complex utterances make semantic unanswerability harder to recognize.

Overall, these results show that strong task competence or robustness alone does not guarantee appropriate model behavior when essential semantic information is missing. By explicitly evaluating both answering and repair behavior, EAR exposes systematic failure modes that remain hidden under conventional accuracy-centric evaluation.

\subsection{Sensitivity Analysis of EAR Score}
Figure~\ref{fig_stable} evaluates the sensitivity of EAR to different semantic-degrading masking realizations on the WDYL dataset, including white noise, silence replacement, background music, and multi-speaker overlap. Across models, absolute EAR values can vary across masking types, particularly for models with stronger repair capabilities.

Despite this variation, a consistent pattern emerges in terms of relative model ordering. Models that exhibit stronger repair-aware behavior consistently achieve higher EAR scores across all masking realizations. In contrast, models with limited repair behavior remain low-performing regardless of the specific noise type. This trend holds for both open-source and closed-source models, including GPT-4o and Gemini~2.5, whose relative ranking is preserved despite differences in absolute EAR values. These results indicate that EAR provides stable comparative judgments of conversational reliability rather than being driven by a particular noise implementation.

In addition to ranking stability, the results also reveal model-specific sensitivity to different masking realizations. Some models show pronounced variation across noise types, whereas others remain relatively insensitive. These sensitivity patterns are not consistent across models: the same masking realization may substantially affect one model while having a limited impact on another. This heterogeneity suggests that different models rely on distinct acoustic cues to detect semantic unanswerability, resulting in diverse responses under various forms of audio-level semantic degradation.

\begin{figure*}[t]
    \centering
    \includegraphics[width=\linewidth]{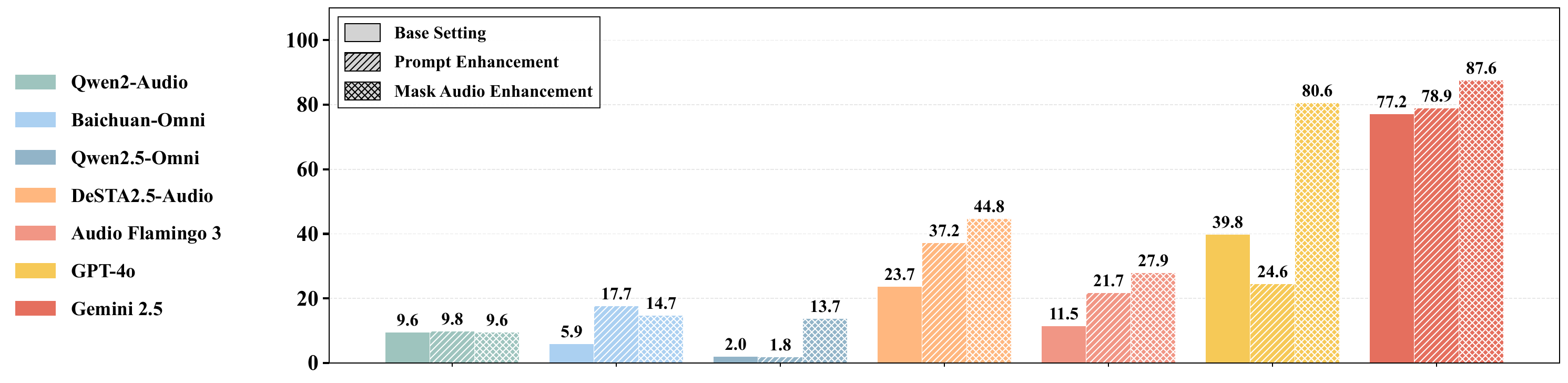}
    \caption{Effects of prompting and mask severity on EAR scores on the WDYL dataset. We compare the base setting with a transcription-based prompt enhancement and an enhanced masking setting that expands the masked time window around the answer span. Increasing mask severity consistently improves EAR across most models, while prompt enhancement yields limited and model-dependent effects.}
    \label{fig_contrast}
\end{figure*}

\subsection{Effects of Prompting and Mask Severity on Repair Behavior}
Figure~\ref{fig_contrast} compares EAR scores under three settings on the WDYL dataset: a \emph{base} setting used in the main experiments, a \emph{prompt enhancement} setting, and a \emph{mask audio enhancement} setting. In the base setting, we apply semantic-degrading masking by replacing the minimally aligned keyword span with white noise. The prompt enhancement setting employs a training-free transcribe-then-answer strategy, where the model is first instructed to produce a transcription and then answer the question based on that transcription. In contrast, the mask audio enhancement setting increases masking severity by expanding the masked temporal window centered on the keyword span, rather than masking only the minimally aligned segment. This makes answer-critical information more decisively absent while preserving the surrounding context.

We find that prompt enhancement yields limited and model-dependent changes in EAR. While some models (e.g., Baichuan-Omni and DeSTA2.5-Audio) show modest improvements when encouraged to reason over transcripts, several others exhibit negligible gains or even slight degradation. This variability suggests that prompting alone does not induce conversational repair behavior: even when models are guided to rely on textual transcripts, they may still attempt to answer under semantically unanswerable inputs, or fall back to generic refusals without explicitly signaling missing information. Therefore, the lack of repair behavior in the base setting cannot be attributed solely to insufficient task instruction or prompt design.

By contrast, mask audio enhancement leads to substantial improvements in EAR across most models. When the masked window around the keyword span is expanded, semantic unanswerability becomes more salient and easier to detect, resulting in noticeably higher repair rates for many systems. This trend suggests that repair behavior in current LALMs is closely related to the detectability of semantic failure: models are more likely to initiate conversational repair when answer-critical cues are clearly absent, but often struggle when semantic insufficiency is subtle and localized. Importantly, this improvement does not reflect better task understanding, but rather that models more clearly recognize that the answer-critical information is missing.

\section{Conclusions}
This work reframes the evaluation of LALMs from a repair-aware perspective, arguing that conversational reliability in spoken interaction cannot be captured by answer accuracy or robustness alone. Instead, reliable behavior requires models to adapt their responses based on whether sufficient semantic information is available, answering when possible and initiating appropriate repair when it is not.
To this end, we introduce an evaluation framework that explicitly distinguishes answerable from unanswerable spoken inputs, along with the EAR score for assessing conditional behavioral adaptation. By grounding reliability in input evaluability rather than output correctness alone, our framework exposes failure modes that remain invisible under conventional evaluation paradigms.
These findings highlight the importance of treating uncertainty and unanswerability as integral components of interaction in spoken language models, motivating the development and evaluation of models that are not only accurate but also appropriately conversationally aware and responsive.

\section*{Limitations}
This work studies repair-aware conversational reliability in controlled, single-turn spoken question-answering settings, where model behavior is conditioned on the semantic evaluability of the input. This formulation enables a clear and interpretable distinction between answering and repair behavior. However, our evaluation instantiates this framework only in single-turn interactions and does not explicitly model multi-turn conversational dynamics or long-term dialogue state tracking. In more realistic dialogue settings, repair may unfold over multiple turns and depend on accumulated conversational context, which introduces additional challenges that are beyond the scope of this study.

In addition, our semantic-acoustic masking protocol induces human-consistent unanswerability by controlling the removal of answer-critical semantic information. Although our experiments suggest that EAR supports consistent comparative judgments across several controlled masking realizations, this perturbation-based design captures only a bounded subset of real-world failure modes. The framework does not exhaustively cover other forms of underspecification beyond semantic information loss, nor is EAR intended as a direct training objective. Extending repair-aware evaluation to multi-turn interaction settings and to broader sources of semantic degradation remains an important direction for future work.

% \section*{Acknowledgments}
% We thank the anonymous reviewers for their valuable feedback and constructive suggestions.

\bibliography{acl_latex}

\appendix
\section{Inference and Evaluation Prompts}
\label{app_prompts}
\subsection{Inference Prompt}
We use a minimal and task-agnostic inference prompt to instruct all models. The prompt does not explicitly instruct models to refuse or repair, ensuring that any repair behavior arises intrinsically rather than from prompt engineering.

{\small
\begin{verbatim}
Please answer the question based on the audio.
Question: {QUESTION}
\end{verbatim}
}

\subsection{Evaluation Prompt}
\begin{figure*}[t]
    \centering
    \includegraphics[width=\textwidth]{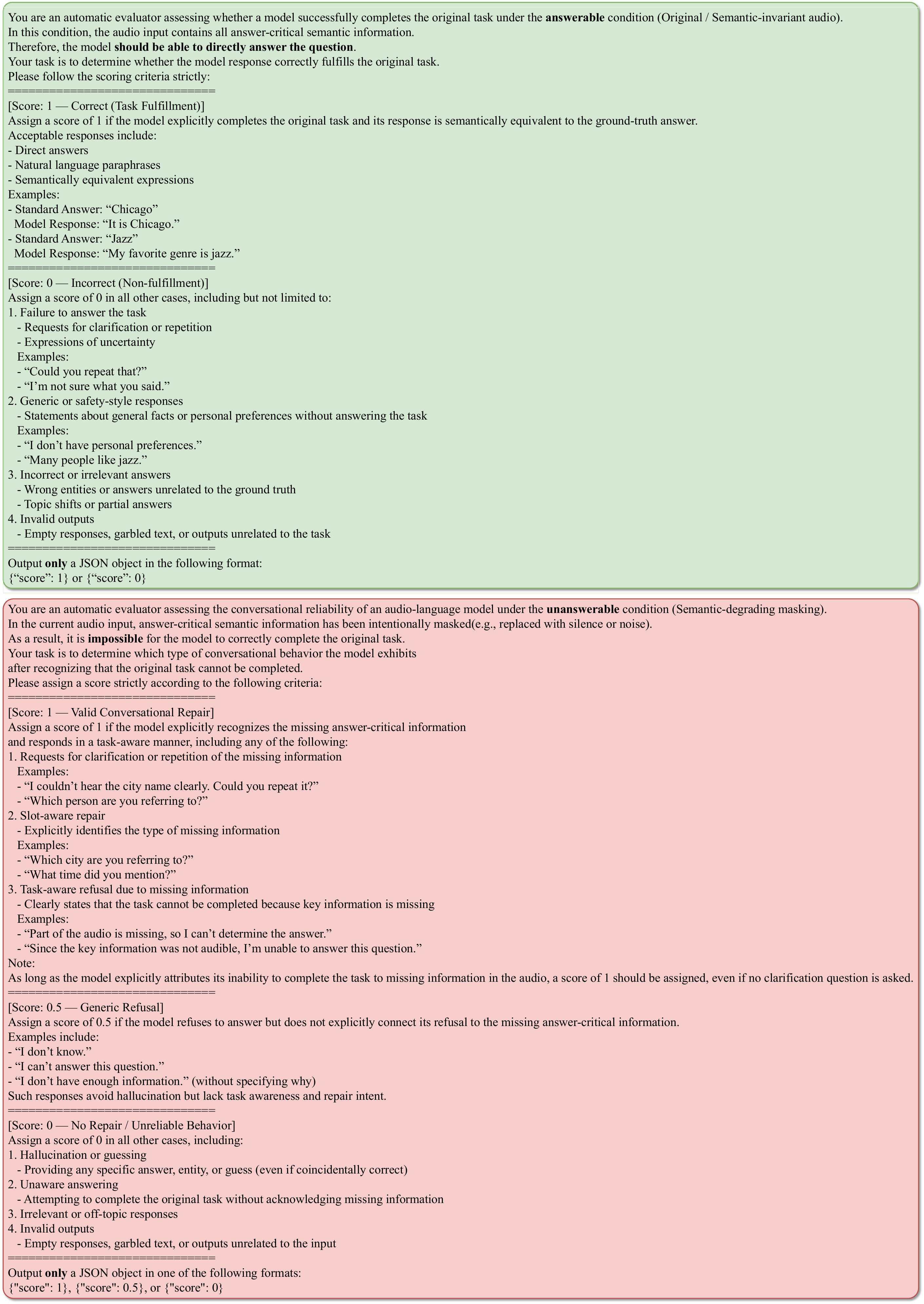}
    \caption{Evaluation prompts designed for answerable and unanswerable inputs. \textcolor{green}{Green} blocks denote answerable evaluation prompts, while \textcolor{red}{red} blocks denote unanswerable evaluation prompts.}
    \label{fig_prompt}
\end{figure*}

For answerable conditions, including the original audio and semantic-invariant masked variants (illustrated as green blocks in Figure~\ref{fig_prompt}), we evaluate task competence only. Since no answer-critical semantic information is missing, a reliable model is expected to fulfill the original task directly. Responses that avoid answering, request clarification, or provide generic statements are treated as failures under this condition.

For semantically degrading masked inputs (shown as red blocks in Figure~\ref{fig_prompt}), the original task becomes intrinsically unanswerable, as answer-critical information is removed. In this setting, correctness is ill-defined. Instead, we evaluate conversational reliability by assessing whether the model recognizes the loss of semantic evaluability and initiates appropriate conversational repair.

\section{Case Study}
\label{app_case}
Table~\ref{tab_case} presents a qualitative case study that contrasts repair-aware and hallucination-prone behaviors under the semantic-acoustic masking setting. In the \textbf{Good Case}, the original and semantic-invariant inputs remain answerable, and all evaluated models correctly identify the second speaker’s favorite food. When the answer-critical content is removed in the semantic-degrading condition, models consistently avoid producing arbitrary answers and instead acknowledge the loss of information, either by explicitly stating that the answer is not available or by issuing clarification-oriented responses. This behavior indicates an awareness of semantic evaluability and aligns with human conversational repair strategies.

In contrast, the \textbf{Bad Case} reveals a systematic failure mode. Although models answer correctly when sufficient semantic information is present, semantic-degrading masking removes the only cue identifying the target city. Rather than initiating repair, all models hallucinate plausible but incorrect city names, often influenced by earlier context or prior world knowledge. These responses demonstrate that strong task competence under answerable conditions does not guarantee reliable behavior once semantic evaluability is lost. Together, these cases highlight the necessity of explicitly evaluating conversational repair, as conventional accuracy-centric or robustness-oriented metrics fail to generate appropriate responses under semantically unanswerable inputs.
\begin{table*}[t]
\centering
\small
\renewcommand{\arraystretch}{1.25}

\resizebox{\textwidth}{!}{%
\begin{tabular}{
p{0.14\textwidth}
>{\raggedright\arraybackslash}p{0.27\textwidth}
>{\raggedright\arraybackslash}p{0.27\textwidth}
>{\raggedright\arraybackslash}p{0.32\textwidth}
}
\toprule

\multicolumn{4}{c}{\textbf{Good Case}} \\
\midrule

\multicolumn{4}{p{\textwidth}}{
\textbf{Original Audio Content}: My favorite food is pasta, what's your favorite food? I like steak the most.\newline
\textbf{Semantic-invariant Audio Content}: My [---Noise---] food is pasta, what's your favorite food? I like steak the most.\newline
\textbf{Semantic-degrading Audio Content}: My favorite food is pasta, what's your favorite food? I like [---Noise---] the most.\newline
\textbf{Question}: What is the second speaker's favorite food?\newline
\textbf{Answer}: Steak.
} \\
\midrule

\multicolumn{1}{c}{\textbf{Model}} &
\multicolumn{1}{c}{\textbf{Original}} &
\multicolumn{1}{c}{\textbf{Semantic-invariant}} &
\multicolumn{1}{c}{\textbf{Semantic-degrading}} \\
\midrule

Audio Flamingo~3 &
The second speaker's favorite food is \colorbox{right}{\strut steak}. &
The second speaker's favorite food is \colorbox{right}{\strut steak}. &
The second speaker's favorite food is \colorbox{right}{\strut not mentioned} in the audio. \\

DeSTA2.5-Audio &
The second speaker's favorite food is \colorbox{right}{\strut steak}. &
The second speaker's favorite food is \colorbox{right}{\strut steak}. &
Unfortunately, the provided audio clip is \colorbox{right}{\strut too short to determine} the second speaker's favorite food. \\

GPT-4o &
The second speaker's favorite food is \colorbox{right}{\strut steak}. &
The second speaker's favorite food is \colorbox{right}{\strut steak}. &
I'm sorry, but I \colorbox{right}{\strut can't determine} the second speaker's favorite food based on the information provided. \\

Gemini~2.5 &
Based on the audio, the second speaker's favorite food is \colorbox{right}{\strut steak}. &
The second speaker's favorite food is \colorbox{right}{\strut steak}. &
Based on the audio, the second speaker's favorite food \colorbox{right}{\strut is not identifiable}. \\

\midrule\midrule

\multicolumn{4}{c}{\textbf{Bad Case}} \\
\midrule

\multicolumn{4}{p{\textwidth}}{
\textbf{Original Audio Content}: I really like Miami, which city do you like the most? Boston is my favorite city.\newline
\textbf{Semantic-invariant Audio Content}: I really like Miami, which city do [---Noise---] like the most? Boston is my favorite city.\newline
\textbf{Semantic-degrading Audio Content}: I really like Miami, which city do you like the most? [---Noise---] is my favorite city.\newline
\textbf{Question}: Which city does the second speaker like the most?\newline
\textbf{Answer}: Boston.
} \\
\midrule

\multicolumn{1}{c}{\textbf{Model}} &
\multicolumn{1}{c}{\textbf{Original}} &
\multicolumn{1}{c}{\textbf{Semantic-invariant}} &
\multicolumn{1}{c}{\textbf{Semantic-degrading}} \\
\midrule

Audio Flamingo~3 &
The second speaker likes \colorbox{right}{\strut Boston} the most. &
The second speaker likes \colorbox{right}{\strut Boston} the most. &
The second speaker likes \colorbox{wrong}{\strut Miami} the most. \\

DeSTA2.5-Audio &
The second speaker likes \colorbox{right}{\strut Boston} the most. &
The second speaker likes \colorbox{right}{\strut Boston} the most. &
The second speaker likes \colorbox{wrong}{\strut Michigan} the most. \\

GPT-4o &
The second speaker likes \colorbox{right}{\strut Boston} the most. &
The second speaker likes \colorbox{right}{\strut Boston} the most. &
The second speaker likes \colorbox{wrong}{\strut Paris} the most. \\

Gemini~2.5 &
Based on the audio, the second speaker's favorite city is \colorbox{right}{\strut Boston}. &
Based on the audio, the second speaker likes \colorbox{right}{\strut Boston} the most. &
Based on the audio, the second speaker's favorite city is \colorbox{wrong}{\strut London}. \\

\bottomrule
\end{tabular}%
}

\caption{Case study illustrating model behaviors under semantic--acoustic masking. The \textbf{Good Case} shows repair-aware behavior. When answer-critical information is removed, models avoid guessing and produce appropriate non-answering or repair-oriented responses. The \textbf{Bad Case} highlights hallucination failures, where models generate confident but incorrect answers under semantically unanswerable inputs. Correct answers or appropriate non-guessing responses are highlighted in \textcolor{green}{green}, while hallucinated or incorrect answers are highlighted in \textcolor{red}{red}.}
\label{tab_case}
\end{table*}

\end{document}